\documentclass[preprint,10pt,3p]{elsarticle}
\usepackage{geometry}
\usepackage{booktabs}
\usepackage[table]{xcolor}
\usepackage{tabularx}
\usepackage{graphicx}
\usepackage{multirow}
\usepackage{caption}
\usepackage{amsmath}
\usepackage{amssymb}
\usepackage{algorithm}
\usepackage{algpseudocode}
\usepackage{pdfpages}

\usepackage{tcolorbox}

\usepackage{amsfonts}
\usepackage{subfigure}
\usepackage{times}
\usepackage{enumitem}
\usepackage{amsthm}
\usepackage{todonotes}
\usepackage[colorlinks=false]{hyperref}

\begin{document}

\begin{frontmatter}

\title{Large Language Model-Assisted Preparation of Transportation Management Plans: A Case Study with WisDOT WisTMP System}

\author[1]{Zihao Sheng}
\author[2]{Pei Li}
\author[1]{Zilin Huang}
\author[3]{Yen-Jung Chen}
\author[1]{Yuhao Luo}
\author[1]{Zhengyang Wan}
\author[1]{Steven T. Parker}
\author[1]{David A. Noyce}
\author[1]{Sikai Chen\corref{cor}}

\cortext[cor]{Corresponding author: Sikai Chen (\texttt{sikai.chen@wisc.edu}).}

\address[1]{Department of Civil and Environmental Engineering, University of Wisconsin-Madison, Madison, WI 53706, USA}
\address[2]{Department of Civil and Architectural Engineering and Construction Management, University of Wyoming, WY 82071, USA} 
\address[3]{Elmore Family School of Electrical and Computer Engineering, Purdue University, West Lafayette, IN 47907, USA} 

\begin{abstract}
Work zones are critical yet hazardous components of transportation infrastructure, requiring carefully designed Transportation Management Plans (TMPs) to ensure safety and mobility. However, TMP preparation remains labor-intensive and heavily dependent on practitioner expertise. This paper proposes a domain-specific Large Language Model (LLM)-assisted framework to investigate the potential of locally deployable open-source LLMs for TMP content generation, leveraging the WisDOT WisTMP system as the application context. The framework fine-tunes multiple open-source LLMs across different model scales and deploys them locally to ensure data security. To support model training, we construct a domain-specific dataset from historical WisTMP documents by converting PDF files into structured question–answer pairs in JSON format. Experimental results show that fine-tuning significantly improves performance across standard text generation metrics and outperforms few-shot prompting. Further section-wise and strategy-level analyses reveal that, while LLMs achieve strong overall performance, they tend to over-generate strategies and struggle to produce project-specific justifications and accurate cost estimates. In addition, scaling from 7B/8B to 14B yields limited gains. These findings demonstrate the potential of LLMs to improve TMP preparation efficiency while highlighting remaining challenges in LLM-assisted TMP development. The source code and demo videos are publicly available at \href{https://zihaosheng.github.io/TMP-LLM/}{\textcolor{blue}{https://zihaosheng.github.io/TMP-LLM/}}.

\end{abstract}


\end{frontmatter}

\section{Introduction}

Work zones are an essential but hazardous component of transportation infrastructure maintenance and improvement~\citep{cheng2022work,liu2025impact}. As the nation's roadway network ages and traffic volumes continue to grow, the frequency and scale of construction and maintenance activities have increased substantially. However, work zones inherently disrupt normal traffic flow, creating safety risks for both motorists and construction workers. According to~\cite{workzone2023}, U.S. work zone crashes in 2023 resulted in 899 fatalities, approximately 39,000 injuries, and about 101,000 total incidents nationwide. In Wisconsin, there are on average 12 fatalities per year in work zones, with drivers and passengers making up a majority of those killed and injured~\citep{wisconsin_fdm}. Speed, inattention, and impaired driving remain the leading contributing factors in work zone crashes. These alarming statistics highlight the critical need for comprehensive and well-designed traffic management strategies during road construction and maintenance activities.

Recognizing the growing safety and mobility challenges posed by work zones, the Federal Highway Administration (FHWA) issued the Final Rule on Work Zone Safety and Mobility in 2004~\citep{fhwa_workzone_facts}. The rule applies to all federally funded highway projects and mandates the development of a Transportation Management Plan (TMP), which outlines coordinated strategies to mitigate work zone impacts~\citep{cottrell2005guidelines,dickerson2016work}. These strategies include temporary traffic control, public information and outreach, and operational measures such as traffic and incident management. The rule adopts a policy-based approach that emphasizes early integration of work zone considerations, systematic impact management, and ongoing performance monitoring, with TMP scope and detail tailored to project-specific impacts.

The Wisconsin Department of Transportation (WisDOT) has implemented the WisTMP system, developed in partnership with the University of Wisconsin TOPS Lab, to facilitate the development, routing, approval, and storage of TMPs for road projects across the state~\citep{uw_transportal_tmp}. The WisTMP system organizes each plan into ten structured sections: (1) Project Information, (2) Project Description, (3) Existing Conditions, (4) Work Zone Strategies, (5) Work Zone Impacts, (6) Traffic Analysis, (7) Public Information Strategies, (8) Incident Management Strategies, (9) Staging Plans, and (10) Additional Information. The first three sections capture the fundamental project context, including basic project metadata, a description of the scope of work, and existing roadway and traffic conditions such as posted speed limits, current capacity, truck percentages, and the presence of queuing. Sections 4 through 10 constitute the operational core of the TMP, covering strategy selection and analytical assessments. In practice, developing these sections requires synthesizing multiple sources of information and translating them into appropriate management strategies, a process that remains largely dependent on practitioner expertise.

Several challenges characterize the current TMP preparation process. First, the selection of appropriate work zone strategies requires engineers to consider a wide range of factors, including roadway geometry, traffic volumes, project duration, detour availability, and the presence of sensitive land uses such as schools and hospitals. The need to integrate these diverse factors creates substantial cognitive complexity and often leads to variability in the quality and completeness of TMPs across projects. Second, TMP preparation is typically performed with limited decision-support tools, requiring engineers to rely heavily on institutional knowledge rather than systematic, data-driven methods. 
Third, TMP development often involves multiple rounds of review and revision, where senior engineers must carefully evaluate and provide feedback on drafts prepared by junior engineers. This iterative process requires significant communication and coordination, and can take several days or even weeks to complete, which introduces significant delays into the project timeline. 
Finally, the volume of infrastructure projects continues to grow, particularly under the Bipartisan Infrastructure Law, which provides unprecedented federal funding for transportation improvements. As a result, the demand for TMP preparation is increasing, further straining the capacity of transportation agencies. 
These challenges collectively highlight the need for intelligent, automated tools to support more efficient, consistent, and scalable TMP development.

Recent advances in large language models (LLMs) such as GPT-4~\citep{achiam2023gpt}, Qwen~\citep{yang2025qwen3}, and DeepSeek~\citep{guo2025deepseek} have demonstrated remarkable capabilities in text understanding, generation, and reasoning across a wide range of tasks. Pre-trained on massive text corpora and further adaptable through fine-tuning or retrieval-augmented generation (RAG), these models can be effectively customized for specialized applications with relatively modest computational resources~\citep{mao2025survey,zhao2026retrieval}. In the transportation domain, a growing body of research has explored LLM applications including traffic prediction~\citep{guo2024towards}, autonomous driving~\citep{huang2025vlm,qu2026found}, crash analysis~\citep{zhao2025safetraffic,sheng2025safeplug}, and traffic scenario generation~\citep {li2024chatsumo,sheng2025talk2traffic}, demonstrating that LLMs can effectively process large-scale, multimodal datasets and provide enhanced analytical capabilities compared to traditional methods. However, the potential of LLMs to assist in the preparation of planning and regulatory documents, such as Transportation Management Plans, remains largely unexplored. This gap is particularly notable given the labor-intensive nature and structured, text-intensive format of TMPs, which make them well-suited for LLM-based assistance.

To address this gap, this paper proposes a domain-specific LLM-assisted framework for TMP preparation, aiming to investigate the potential, limitations, and practical feasibility of applying locally deployable open-source LLMs to this specialized transportation engineering task. Considering the structured nature of TMP documents, the framework formulates TMP preparation as a set of section-specific generation tasks, in which the main operational sections (Sections 4–9) are generated based on the project information provided in the foundational sections (Sections 1–3). Historical TMP documents are transformed into a domain-specific instruction-tuning dataset to adapt LLMs to the terminology, document structure, and planning practices of the WisTMP system.
Experimental results demonstrate that fine-tuning yields substantial performance improvements across all evaluated models, as measured by standard text generation metrics including BLEU-4, ROUGE-1, ROUGE-2, and ROUGE-L. Qwen2.5-7B-Instruct-1M achieves the highest overall average score, while scaling from 7B/8B to 14B yields only marginal gains. Beyond text-level metrics, section-wise analysis reveals variation in performance across different TMP components. At the strategy level, fine-tuned models achieve reasonable recall but tend to over-generate, highlighting the challenges in complex, domain-specific aspects of TMP preparation.
The main contributions of this study are summarized as follows:
\begin{itemize}
    \item \textbf{Domain-specific dataset construction and task formulation.} We develop a document processing and data construction pipeline tailored to the structural characteristics of WisTMP documents, including cross-page table merging, version filtering, data sanitization, and section-based QA pair construction. The framework formulates TMP preparation as section-specific generation tasks that map foundational project information to operational planning content, which enables LLMs to learn from historical TMP preparation practices.
    
    \item \textbf{Systematic evaluation and domain-specific insights.}  We systematically investigate the effectiveness and limitations of multiple open-source LLMs across different model families and parameter scales for TMP content generation. Beyond standard text metrics, we evaluate strategy-level precision and recall against the standardized WisTMP strategy list, compare fine-tuning with few-shot prompting, and analyze errors in justifications and cost estimates. These analyses show which parts of a TMP can be generated reliably and which still need practitioner input.
    
    \item \textbf{Local deployment.} Considering the data privacy constraints associated with transportation infrastructure projects, we deploy the fine-tuned open-source LLMs on local institutional hardware and develop an interactive prototype for TMP content generation. This provides a practical foundation for integrating LLM-assisted drafting into existing TMP preparation workflows without transmitting sensitive project information to external LLM service providers.
\end{itemize}

\section{Related Works}

\subsection{Work Zone Management and TMP Practices}

Transportation Management Plans (TMPs) are essential for mitigating safety and mobility impacts in work zones, yet their development remains a labor-intensive process~\citep{uw_transportal_tmp,cottrell2005guidelines,dickerson2016work}. Engineers must synthesize project characteristics, traffic conditions, roadway geometry, staging constraints, detour feasibility, and stakeholder needs when selecting appropriate work zone strategies and documenting them in a consistent form. Although agencies have developed standardized TMP procedures and digital systems to support this process, much of the section-level drafting work still depends on practitioner experience.

Prior work in this area has mainly focused on analytical support for work zone planning rather than TMP text preparation itself~\citep{park2018apriori}. Many studies and technical reports examine lane closure analysis, work zone traffic impacts, mobility assessment, safety evaluation, and simulation-based comparison of alternative control strategies~\citep{racha2008analysis, abdelmohsen2018optimizing, morgado2014work}. These efforts help agencies evaluate candidate treatments and justify strategy choices, but their outputs are typically performance measures, impact estimates, or scenario comparisons rather than draft TMP narratives. Other efforts emphasize standardized procedures, strategy selection frameworks, and digital decision-support systems for work zone planning and TMP-related management~\citep{boodlal2020strategies, adeli2004intelligent, zhou2015simplified}. While these tools improve process consistency, they provide limited support for transforming structured project information into section-specific TMP content.

Overall, existing work has been more successful in supporting strategy analysis and workflow standardization than document generation. This leaves a practical gap between historical TMP knowledge and the efficient drafting of project-specific TMP sections, which motivates the use of data-driven language models for TMP preparation.

\subsection{LLMs in Transportation}
Large language models (LLMs) have recently been applied to a broad range of transportation tasks, driven by their strong capabilities in natural language understanding, contextual reasoning, and knowledge synthesis~\citep{jin2026survey,jiang2025survey,zhang2026large,huang2026sim2real}.

One prominent line of research focuses on traffic analysis and management. LLMs have been used for traffic flow forecasting by encoding spatiotemporal patterns into natural language representations~\citep{guo2024towards,rong2024large,xu2025gpt4tfp}. Beyond prediction, researchers have also leveraged LLMs for traffic signal control and network-level traffic management, where the models serve as decision-making agents that interpret real-time conditions and generate control strategies~\citep{movahedi2024crossroads,fu2026llm}.

Another active area is autonomous driving, where LLMs have been integrated into multiple modules of the driving systems for tasks such as scene understanding, trajectory planning, and natural language-based human-vehicle interaction~\citep{shaw2025saferoute,qu2025vl,huang2025sky,sheng2026explorevla}. More recently, LLMs have also been used to generate safety-critical driving scenarios for simulation-based testing and to work in conjunction with reinforcement learning frameworks for training autonomous vehicle policies~\citep{zhang2024chatscene,sheng2026curricuvlm,huang2026drivevlm}.
These efforts highlight the potential of LLMs to bridge the gap between unstructured language inputs and structured driving decisions.

LLMs have also shown promise in transportation safety applications. Prior work has investigated their use in crash severity prediction, safety-critical event detection, and the comprehension of traffic regulations and standards~\citep{huo2025crash,zhen2025crashsage,cai2026driving}. In addition, a growing body of work examines the use of LLMs as assistants for transportation planning and decision support. Examples include travel demand modeling, policy analysis, and public transit service evaluation~\citep{gong2024mobility,shao2025personal,liu2025llm}. These studies suggest that LLMs can serve not only as analytical tools but also as interfaces for accessing and synthesizing domain knowledge.
While these studies demonstrate the versatility of LLMs across diverse transportation applications, the use of LLMs for generating structured, regulation-compliant planning documents has received limited attention. In particular, the drafting of TMPs remains largely unexplored.

\subsection{LLM-Assisted Document and Report Generation}

Building on LLMs' capabilities in language understanding and reasoning, recent research has begun to explore their potential for automating the drafting of complex, domain-specific documents and reports.
In the medical domain, LLMs have been widely adopted to streamline complex clinical documentation~\citep{goyal2024healai,seo2024evaluation,satheakeerthy2025llm}. For example, studies have demonstrated the ability of LLMs to synthesize a series of clinical admission notes into comprehensive hospital discharge summaries~\citep{rodrigues2025large}, and leverage multimodal knowledge graphs to translate X-ray images and disease tokens into structured radiology reports~\citep{wang2025r2genkg}. Similarly, the legal sector has actively embraced LLMs for formal document generation~\citep{zhou2025lawgpt, li2025legalagentbench,siino2025exploring}. Researchers have developed pipelines that harness LLMs to automatically generate executable smart contracts directly from plain-text legal agreements~\citep{napoli2024leveraging}. Furthermore, LLMs are increasingly utilized in the architectural, engineering, and construction (AEC) industry for automated compliance reporting~\citep{kampelopoulos2025review, chen2024automated}. For instance, advanced vision-language agents have been deployed to evaluate site images and textual data against building codes, automatically generating compliance inspection reports that identify specific structural issues and explicitly cite relevant regulations~\citep{zheng2025vision}.
In this work, we extend LLM-assisted document generation 
to the transportation domain by investigating the use of fine-tuned LLMs for 
drafting TMPs.

\section{Methodology}
\label{sec:methodology}

\begin{figure}[t]
    \centering
    \includegraphics[width=0.99\linewidth]{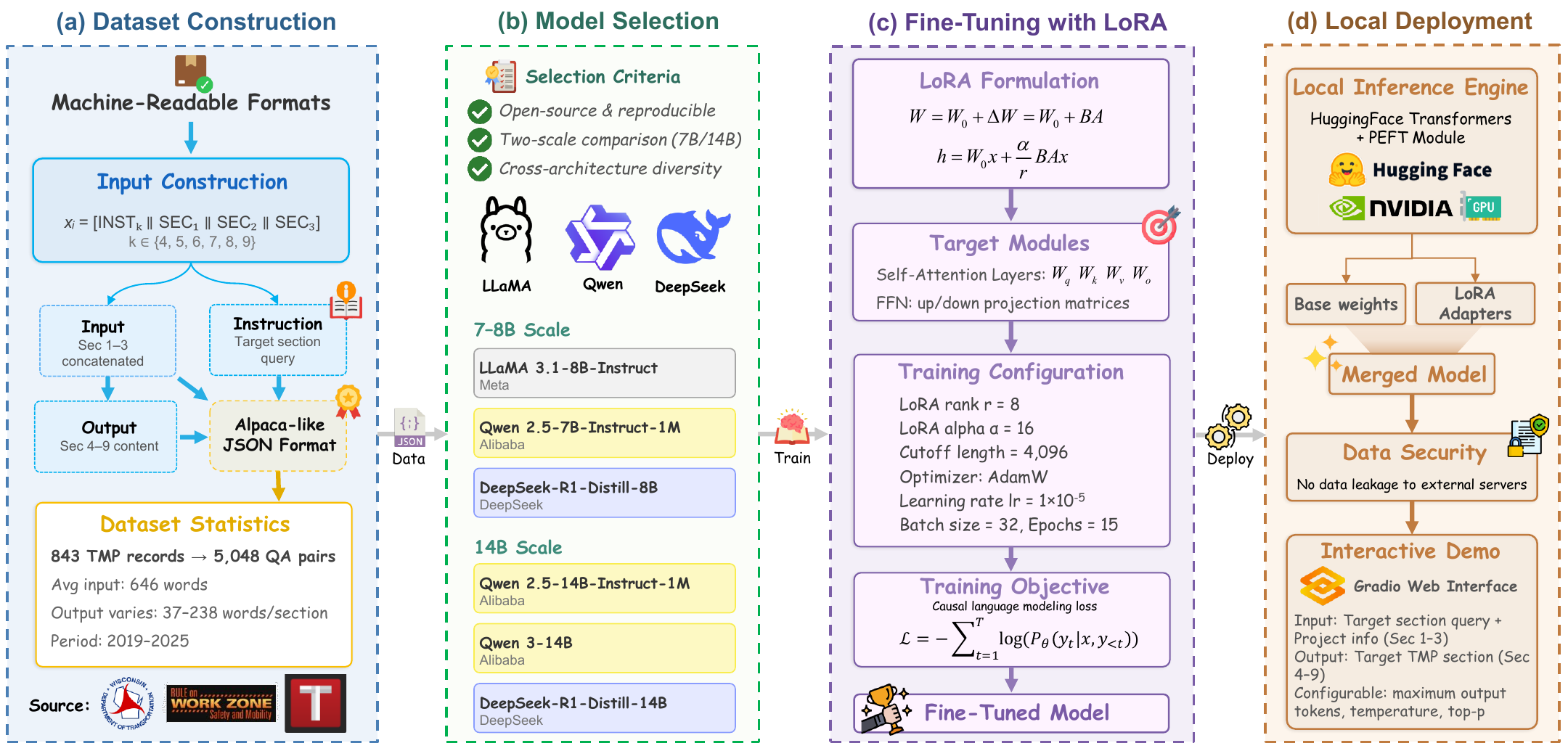}
    \caption{Overview of the proposed LLM-assisted TMP preparation framework, comprising four stages: dataset construction from historical WisTMP records, model selection across three families and two scales, fine-tuning with LoRA, and local deployment with an interactive web interface.}
    \label{fig:methodology}
\end{figure}

This section presents the proposed framework for LLM-assisted TMP preparation. 
The framework is designed around three practical requirements of TMP preparation: learning from structured historical planning documents, generating section-specific content conditioned on project context, and supporting local deployment under data privacy constraints.
As shown in Fig.~\ref{fig:methodology}, the overall workflow consists of four main stages: (1) dataset construction from historical WisTMP documents, (2) model selection across three families and two parameter scales, (3) fine-tuning of open-source LLMs on the constructed dataset, and (4) local deployment of the fine-tuned models for inference.

\subsection{Dataset Construction}
\label{subsec:dataset}

\subsubsection{Data Source}

The dataset used in this study is derived from the WisTMP system. Each document in the repository represents a formal TMP developed for real-world road construction projects, containing structured descriptions of project conditions, anticipated impacts, and corresponding management measures.

In the WisTMP system, each TMP consists of ten structured sections: (1) Project Information, (2) Project Description, (3) Existing Conditions, (4) Work Zone Strategies, (5) Work Zone Impacts, (6) Traffic Analysis, (7) Public Information Strategies, (8) Incident Management Strategies, (9) Staging Plans, and (10) Additional Information.
The first three sections primarily describe the project background, location, and existing operating conditions, while Sections 4 through 9 contain the main operational and planning content of the TMP. Section 10 typically includes supplementary details and is not considered in this study. 
Sections 4 through 9 are typically the most time-consuming to prepare, as they require engineers to integrate diverse project-specific factors and rely on professional judgment to develop appropriate management measures. 
To help readers better understand the structure and content of WisTMP documents, a representative TMP example with sensitive information removed is provided in \ref{appendix:tmp_example}. 

Here we formalize the data structure used for supervised fine-tuning. Let $\mathcal{D} = \{(x_i, y_i)\}_{i=1}^{N}$ denote the training dataset, where each sample consists of an input $x_i$ and a target output $y_i$. The input $x_i$ is constructed as:
\begin{equation}
    x_i = [\texttt{INST}_k \, \| \, \texttt{SEC}_1^{(i)} \, \| \, \texttt{SEC}_2^{(i)} \, \| \, \texttt{SEC}_3^{(i)}]
    \label{eq:input}
\end{equation}
where $\texttt{INST}_k$ denotes the instruction prompt for target section $k \in \{4, 5, 6, 7, 8, 9\}$, $\texttt{SEC}_j^{(i)}$ represents the text content of Section~$j$ from the $i$-th TMP document, and $\|$ denotes string concatenation. The corresponding output $y_i = \texttt{SEC}_k^{(i)}$ is the ground-truth content of target section $k$.
In this study, we utilize WisTMP documents from 2019 to 2025. 
The overall dataset construction pipeline is illustrated in Fig.~\ref{fig:data_pipeline}.

\subsubsection{Data Corpus Extraction and Cleaning}\label{sec:data_corpus}

\begin{figure}[t]
    \centering
    \includegraphics[width=0.89\linewidth]{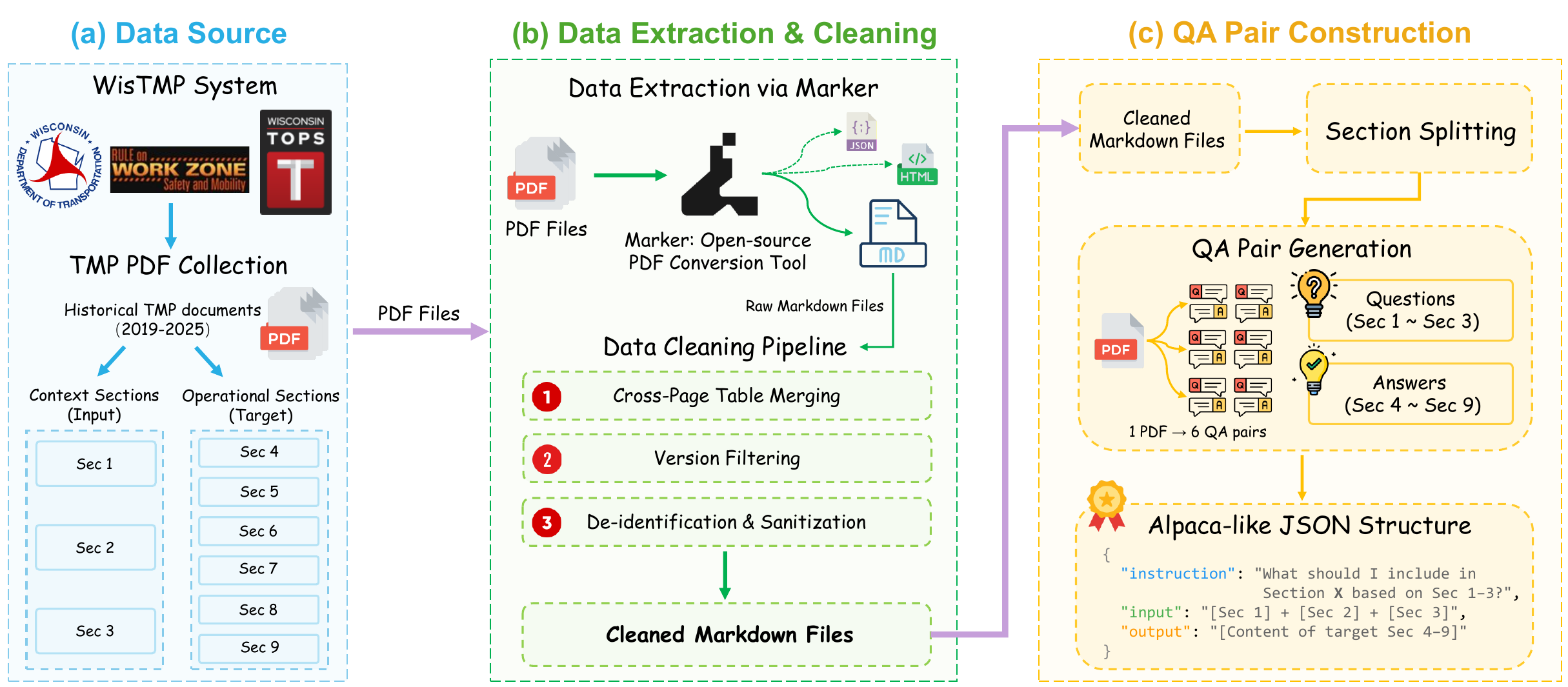}
    \caption{Pipeline of the dataset construction for TMP preparation, including data sourcing from the WisTMP system, PDF-to-Markdown conversion and cleaning, and section-based QA pair construction for supervised fine-tuning.}
    \label{fig:data_pipeline}
\end{figure}

A central challenge in dataset construction is converting TMP documents from raw PDF files into machine-readable formats suitable for model training, as illustrated in Fig.~\ref{fig:data_pipeline}(b). To address this, we use an open-source document processing tool, Marker~\citep{marker_github}, that supports converting PDFs into Markdown. Marker is particularly useful in this context because it preserves complex document elements such as tables and nested structures. 

After converting the PDF documents into Markdown, we apply a multi-stage cleaning pipeline to improve data quality. A major issue encountered during extraction is the handling of tables that span multiple pages. In these cases, Marker may incorrectly split a single table into two separate table blocks. Based on the structural characteristics of TMP documents, we observed that two genuinely independent tables do not typically appear consecutively across a page break without intervening explanatory text or a new section heading. Leveraging this domain-specific observation, we developed a rule-based post-processing step that detects consecutive table fragments and automatically merges them when appropriate.

We then remove unwanted document versions and iterative review content. Because some source files include multiple rounds of revision comments from supervisors or reviewers, we retain only the finalized version of each TMP so that the model learns from completed expert-authored content rather than intermediate drafts. 
In addition, repetitive, non-informative, and potentially identifying elements, such as page numbers, engineer role information, and signatures, are removed from the dataset. This preprocessing step not only reduces noise but also follows the principle of data minimization by removing unnecessary and potentially sensitive information while retaining project-specific information essential for TMP content generation.

\subsubsection{QA Pair Construction}

To support supervised fine-tuning, we transform the cleaned TMP corpus into a collection of question-answer (QA) pairs, as shown in Fig.~\ref{fig:data_pipeline}(c). 
For each QA pair, the input side begins with an instruction specifying the target section to be generated, followed by the concatenated content of Sections 1 through 3, which together provide the essential project context.
The output side consists of one target section selected from Sections 4 through 9. Because the TMP documents are explicitly organized by section headers, a single TMP file can be decomposed into six distinct QA pairs, one for each target section in the operational portion of the document.

After generating these pairs, we format the dataset into an Alpaca-style JSON structure, in which each instance contains three fields: \texttt{instruction}, \texttt{input}, and \texttt{output}. The \texttt{instruction} field specifies the target section to be generated, phrased as a descriptive request such as: ``What should be included in Section X based on the information provided in Sections 1 through 3?'' The \texttt{input} field contains the concatenated content from Sections 1 through 3, providing the contextual information needed for reasoning and generation. The \texttt{output} field contains the full content of the corresponding target section.

This instruction-based QA design ensures that the model receives both comprehensive project context and an explicit generation objective. It also allows a single TMP document to yield multiple training instances, thereby improving data efficiency while preserving the structural relationship between foundational project information and downstream planning content. In this way, the dataset is designed to support context-aware, section-specific generation that is closely aligned with expert TMP preparation practices.
For clarity and reproducibility, the prompt templates used for TMP content generation are provided in \ref{appendix:prompt}.

\begin{figure}[t]
    \centering
    \includegraphics[width=0.99\linewidth]{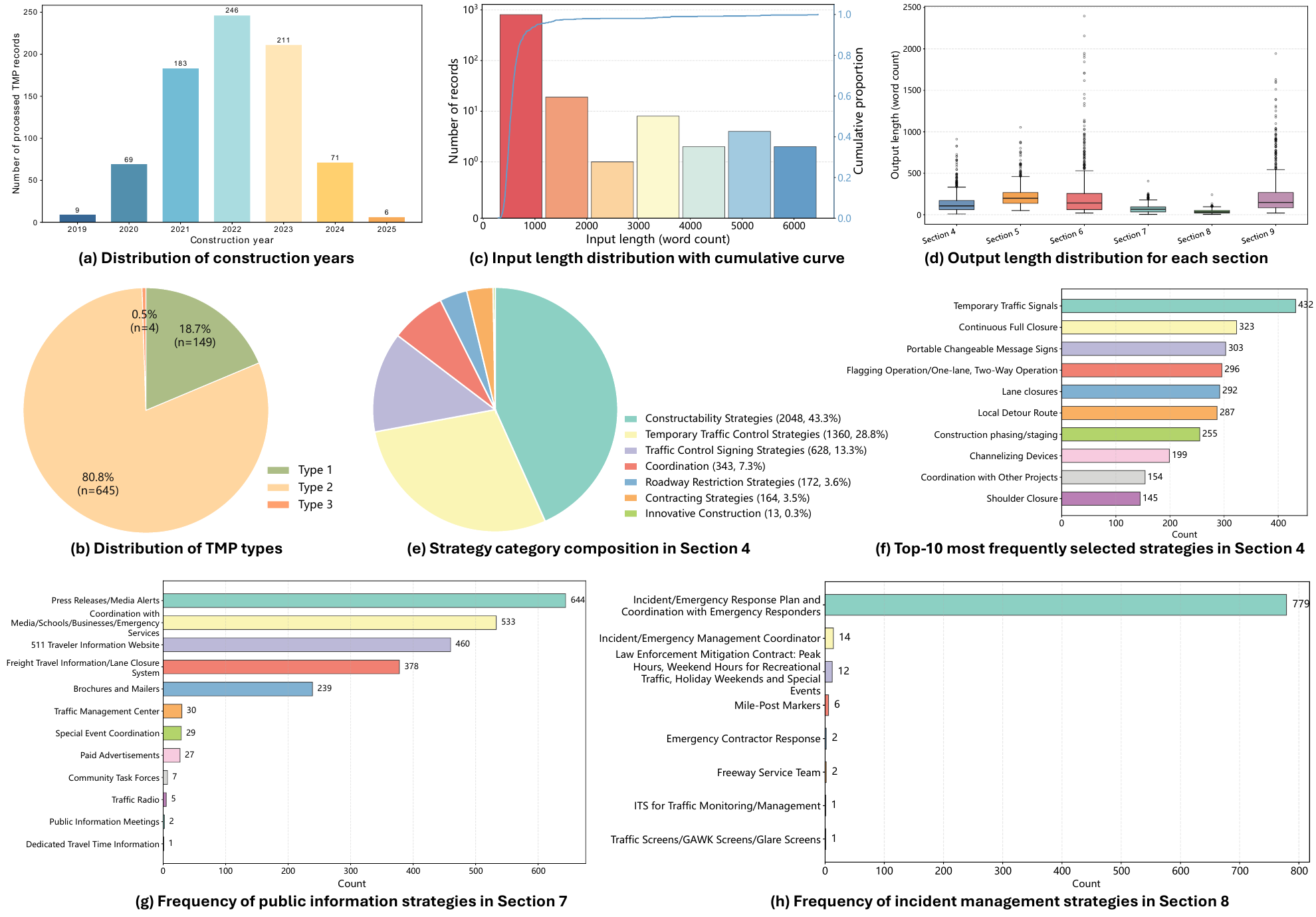}
    \caption{Comprehensive statistics of the curated TMP dataset, including temporal coverage, TMP type composition, input/output length distributions, and strategy distributions across Sections~4, 7, and~8.}
    \label{fig:data_stats}
\end{figure}

\subsubsection{Dataset Statistics}\label{sec:data_stats}
After cleaning and restructuring, the curated corpus comprises 843 processed TMP records.
Each record yields up to six QA pairs, which results in a final dataset of 5,048 instruction–input–output triples. The per-section breakdown is as follows: 835 for Section~4 (Work Zone Strategies), 843 for Section~5 (Work Zone Impacts), 843 for Section~6 (Traffic Analysis), 842 for Section~7 (Public Information Strategies), 842 for Section~8 (Incident Management Strategies), and 843 for Section~9 (Staging Plans).
We characterize the dataset along three dimensions: temporal coverage, TMP type composition, and input–output length distribution, as summarized in Fig.~\ref{fig:data_stats}.

\paragraph{Temporal Coverage}
As shown in Fig.~\ref{fig:data_stats}(a), the majority of processed records fall within the 2021--2023 period, indicating that the corpus primarily captures recent TMP preparation practices. The lower counts observed in 2024 and 2025 are likely due to data incompleteness or reporting lag, as recent projects may not yet be fully recorded.

\paragraph{TMP Type Composition}
As illustrated in Fig.~\ref{fig:data_stats}(b), among the 798 records with a valid TMP type label, Type~2 TMPs constitute the majority (645, 80.8\%), followed by Type~1 (149, 18.7\%) and Type~3 (4, 0.5\%). According to the WisDOT Facilities Development Manual, Type~1 corresponds to low-impact projects, Type~2 to moderate-complexity projects, and Type~3 to large-scale, long-duration projects. The predominance of Type~2 records indicates that the dataset is centered on moderate-complexity planning scenarios.
Accordingly, the experimental findings of this study primarily reflect LLM performance on TMP preparation for low- and moderate-complexity (Type 1 and Type 2) in Wisconsin. The limited representation of Type 3 TMPs prevents a reliable assessment of model performance on highly complex projects.

\paragraph{Input and Output Characteristics}
The distribution of input lengths is presented in Fig.~\ref{fig:data_stats}(c). The concatenated context from Sections~1--3 contains an average of 646 words. 
As shown in Fig.~\ref{fig:data_stats}(d), output lengths vary substantially across target sections: Sections~5, 6, and~9 tend to produce longer outputs (217, 225, and 238 words on average, respectively), while Sections~7 and~8 are considerably shorter (69 and 37 words). Section~4 falls in between at 140 words. 

\paragraph{Strategy Distribution}
To provide further insight into the semantic composition of the dataset, we examine the distribution of strategies selected in Sections~4, 7, and~8. 
As shown in Fig.~\ref{fig:data_stats}(e) and Fig.~\ref{fig:data_stats}(f), the selected strategy categories in Section~4 are concentrated in Constructability Strategies (43.3\%) and Temporary Traffic Control Strategies (28.8\%), with the most frequently occurring individual strategies being continuous full closure, flagging or one-lane two-way operation, and lane closures.
Fig.~\ref{fig:data_stats}(g) shows that Section~7 exhibits a long-tail distribution dominated by a small set of standardized outreach channels such as press releases, coordination with media and emergency services, and traveler information websites. As illustrated in Fig.~\ref{fig:data_stats}(h), Section~8 is the most concentrated, with Incident/Emergency Response Plan and Coordination with Emergency Responders accounting for 95.3\% of all occurrences. These distributional patterns are consistent with the relatively short content observed in Sections~7 and~8 (Fig.~\ref{fig:data_stats}(d)), and suggest that historical TMP practices tend to rely on established strategy templates for public information and incident management.

\subsection{Model Selection}
\label{subsec:models}

We select a diverse set of open-source LLMs to evaluate the effectiveness of fine-tuning for TMP generation.  Table~\ref{tab:models} summarizes the models used in this study. We consider models from three representative families: LLaMA, Qwen, and DeepSeek, spanning two parameter scales (7B/8B and 14B). 

\begin{table}[t]
\centering
\caption{Summary of LLMs selected for fine-tuning.}
\label{tab:models}
\rowcolors{2}{white}{gray!10}
\begin{tabular}{lcc}
\toprule
\rowcolor{gray!30}
\textbf{Model} & \textbf{Parameters} & \textbf{Family} \\
\midrule
meta-llama/Llama-3.1-8B-Instruct & 8B & LLaMA \\
deepseek-ai/DeepSeek-R1-Distill-Llama-8B & 8B & DeepSeek \\
Qwen/Qwen2.5-7B-Instruct-1M & 7B & Qwen \\
\midrule
Qwen/Qwen3-14B & 14B & Qwen \\
deepseek-ai/DeepSeek-R1-Distill-Qwen-14B & 14B & DeepSeek \\
Qwen/Qwen2.5-14B-Instruct-1M & 14B & Qwen \\
\bottomrule
\end{tabular}
\end{table}

\begin{itemize}
    \item \textbf{LLaMA 3.1-8B-Instruct}~\citep{grattafiori2024llama} is an instruction-tuned variant of Meta's LLaMA~3.1 family, pre-trained on a large multilingual corpus and fine-tuned with reinforcement learning from human feedback (RLHF). It serves as a representative baseline from the widely adopted LLaMA ecosystem.

    \item \textbf{Qwen 2.5-7B/14B-Instruct-1M}~\citep{qwen2025qwen25technicalreport} belongs to the Qwen~2.5 series developed by Alibaba. The ``1M'' designation indicates support for context lengths up to one million tokens, which is particularly advantageous for processing long TMP documents containing extensive project descriptions and tabular data. Both the 7B and 14B variants are included to assess the effect of model scale within the same architecture.

    \item \textbf{Qwen3-14B}~\citep{yang2025qwen3} is a more recent release in the Qwen family, incorporating architectural improvements and expanded pre-training data relative to Qwen~2.5. Its inclusion enables a comparison of generational improvements within the same model family at the 14B scale.

    \item \textbf{DeepSeek-R1-Distill-Llama-8B/Qwen-14B}~\citep{guo2025deepseek} are distilled versions of DeepSeek's R1 reasoning model (671B total parameters), which transfers its enhanced reasoning capabilities into more compact architectures. Both the 8B and 14B variants are included to evaluate whether reasoning-oriented pre-training offers advantages for the structured generation task of TMP preparation.
\end{itemize}

The model selection is guided by three considerations.
First, all selected models are open-source and freely available, which enables local deployment without reliance on commercial APIs. Because transportation infrastructure project information is sensitive and subject to data privacy constraints, this study does not consider proprietary LLMs accessed through commercial APIs (e.g., GPT, Claude, and Gemini). In addition, open-source models with fixed checkpoints ensure reproducibility, while proprietary models may be updated without notice.
Second, the two-scale design (7B/8B and 14B) allows us to quantify the trade-off between model size and generation quality. Third, the diversity of model families (LLaMA, Qwen, and DeepSeek) enables cross-architecture comparison and mitigates the risk of architecture-specific biases in our evaluation.

\subsection{Fine-Tuning with Low-Rank Adaptation}
\label{subsec:finetuning}

Fine-tuning all parameters of a multi-billion-parameter LLM is computationally prohibitive for most institutional settings. To enable efficient domain adaptation, we employ Low-Rank Adaptation (LoRA)~\citep{hu2022lora}, a parameter-efficient fine-tuning (PEFT) technique that freezes the pre-trained weights and injects trainable low-rank matrices into the model.

The key insight behind LoRA is that the weight updates during fine-tuning exhibit a low intrinsic rank. Rather than updating the full weight matrix $W_0 \in \mathbb{R}^{d \times k}$ of a pre-trained model, LoRA constrains the update to a low-rank decomposition:
\begin{equation}
    W = W_0 + \Delta W = W_0 + BA
    \label{eq:lora}
\end{equation}
where $B \in \mathbb{R}^{d \times r}$ and $A \in \mathbb{R}^{r \times k}$ are trainable low-rank matrices, and $r \ll \min(d, k)$ is the rank of the decomposition. During fine-tuning, the original weight matrix $W_0$ is frozen, and only the matrices $A$ and $B$ are updated via gradient descent. This reduces the number of trainable parameters from $d \times k$ to $r \times (d + k)$, achieving a compression ratio of approximately $\frac{dk}{r(d+k)}$.
The LoRA update is scaled by a factor $\frac{\alpha}{r}$, where $\alpha$ is a hyperparameter that controls the magnitude of the adaptation, such that the output is:
\begin{equation}
    h = W_0 x + \frac{\alpha}{r} BAx
    \label{eq:lora_forward}
\end{equation}
where $h$ is the output of the linear layer and $x$ is the input. This scaling ensures that the contribution of the LoRA update remains stable across different choices of rank $r$.

In our implementation, LoRA adapters are applied to the query ($W_q$), key ($W_k$), value ($W_v$), and output ($W_o$) projection matrices of the self-attention layers, as well as the up-projection and down-projection matrices of the feed-forward network (FFN) layers in each Transformer block. 
The training objective is the standard causal language modeling loss, which minimizes the negative log-likelihood of the target tokens given the input context:
\begin{equation}
    \mathcal{L} = -\sum_{t=1}^{T} \log P_\theta(y_t \mid x, y_{<t})
    \label{eq:loss}
\end{equation}
where $\theta$ represents the trainable LoRA parameters, $y_t$ is the $t$-th token of the target output, $T$ is the total number of tokens in the output sequence, and $y_{<t}$ denotes the preceding tokens.

\subsection{Local Deployment}
\label{subsec:deployment}

A critical requirement for the proposed framework is data privacy and security. TMP documents may contain sensitive information about transportation infrastructure projects, including specific locations, construction timelines, and detour routes. To mitigate the risk of external data exposure, both model fine-tuning and inference are conducted on local institutional computing infrastructure without relying on external commercial LLM services. Together with the data sanitization procedures described in Section~\ref{sec:data_corpus}, this local computing approach reduces privacy risks throughout the data preparation, model training, and content generation stages. The inference pipeline is implemented using the Hugging Face transformers library with the PEFT module for loading LoRA adapters. At inference time, the merged model (base weights plus LoRA adapters) processes the user-provided project information (Sections 1--3) and generates the requested operational section.

An interactive demonstration system is developed using Gradio~\citep{abid2019gradio}, a Python-based framework for building web-based machine learning interfaces. The system allows users to provide project-specific information (e.g., selected TMP sections) and specify target sections for content generation. Users can also configure key inference parameters, such as maximum output tokens, temperature, and top-$p$ sampling, to control generation behavior. The interface further supports multimodal input, including image uploads, to accommodate future extensions of the system.

\section{Results and Evaluation}

\subsection{Experimental Setup}

\paragraph{Training Configuration} 
The fine-tuning process is conducted using the following configuration. All models are trained on 2$\times$ NVIDIA A100 GPUs. 
The dataset is split into training and test sets at the TMP document level with a ratio of 9:1. All QA pairs derived from the same TMP document are assigned exclusively to either the training set or the test set to prevent data leakage arising from shared project information. 
The LoRA rank is set to $r = 8$, with a scaling factor of $\alpha = 16$. The maximum sequence length (cutoff length) is set to 4{,}096 tokens to accommodate the input lengths of TMP documents. The AdamW optimizer is employed with a learning rate of $1 \times 10^{-5}$. The effective batch size is set to 32 through per-device batch size and gradient accumulation. Training is conducted for 15 epochs on the training set.

\paragraph{Evaluation Metrics}
The quality of the generated TMP content is evaluated at both the text and strategy levels. While conventional text generation metrics measure the similarity between generated and historical TMP content, they may not adequately capture whether individual traffic management strategies are correctly recovered. Therefore, we complement text similarity metrics with domain-specific strategy-level evaluation.

\textbf{BLEU-4}~\citep{papineni2002bleu} (Bilingual Evaluation Understudy) measures the precision of $n$-gram overlaps between the generated and reference texts, with $n$ up to 4. It is computed as:
\begin{equation}
    \text{BLEU-4} = \text{BP} \cdot \exp\left(\sum_{n=1}^{4} w_n \log p_n\right)
    \label{eq:bleu}
\end{equation}
where $p_n$ is the modified $n$-gram precision, $w_n = 1/4$ is the uniform weight for each $n$-gram order, and $\text{BP}$ is a brevity penalty that penalizes overly short outputs.

\textbf{ROUGE-1} and \textbf{ROUGE-2}~\citep{lin2004rouge} (Recall-Oriented Understudy for Gisting Evaluation) measure the recall of unigram and bigram overlaps, respectively, between the generated and reference texts:
\begin{equation}
    \text{ROUGE-}n = \frac{\sum_{s \in \text{Ref}} \sum_{\text{gram}_n \in s} \text{Count}_{\text{match}}(\text{gram}_n)}{\sum_{s \in \text{Ref}} \sum_{\text{gram}_n \in s} \text{Count}(\text{gram}_n)}
    \label{eq:rouge}
\end{equation}
where $\text{Count}_{\text{match}}$ is the number of $n$-grams in the reference that also appear in the generated text.

\textbf{ROUGE-L} measures the longest common subsequence (LCS) between the generated and reference texts, capturing sentence-level structural similarity:
\begin{equation}
    \text{ROUGE-L} = \frac{(1 + \beta^2) \cdot R_{\text{LCS}} \cdot P_{\text{LCS}}}{R_{\text{LCS}} + \beta^2 \cdot P_{\text{LCS}}}
    \label{eq:rouge_l}
\end{equation}
where $R_{\text{LCS}}$ and $P_{\text{LCS}}$ are the LCS-based recall and precision, respectively, and $\beta$ is a parameter that controls the relative importance of recall versus precision.

In addition to text-based metrics, we evaluate strategy-level alignment between generated and ground-truth TMP content. Specifically, we evaluate the strategy \textbf{Recall} and \textbf{Precision} for Sections~4 (Work Zone Strategies), 7 (Public Information Strategies), and 8 (Incident Management Strategies). Recall measures the proportion of strategies selected in the historical TMP that are recovered by the model, while precision measures the proportion of generated strategies that match those selected in the historical TMP:
\begin{equation}
    \text{Recall} = \frac{|\mathcal{S}_{\text{gen}} \cap \mathcal{S}_{\text{ref}}|}{|\mathcal{S}_{\text{ref}}|}, \quad \text{Precision} = \frac{|\mathcal{S}_{\text{gen}} \cap \mathcal{S}_{\text{ref}}|}{|\mathcal{S}_{\text{gen}}|}
    \label{eq:recall_precision}
\end{equation}
where \(S_{\mathrm{gen}}\) and \(S_{\mathrm{ref}}\) denote the sets of strategies in the generated and reference outputs, respectively. These metrics provide domain-specific measures of strategy-level agreement, which enables the identification of strategy omissions and over-generation that may not be adequately reflected by text similarity metrics.

\begin{table}[t]
\centering
\caption{Performance comparison of different LLMs under zero-shot prompting (before fine-tuning), few-shot prompting, and after fine-tuning across BLEU-4 and ROUGE metrics. Few-shot prompting is evaluated on Qwen2.5-7B-Instruct-1M. The best results are highlighted in \textbf{bold}, and the second-best results are \underline{underlined}.}
\label{tab:overall_performance}
\rowcolors{3}{gray!10}{white}
\begin{tabular}{lccccc}
    \toprule
    \textbf{Model} & \textbf{BLEU-4} $\uparrow$ & \textbf{ROUGE-1} $\uparrow$ & \textbf{ROUGE-2} $\uparrow$ & \textbf{ROUGE-L} $\uparrow$ & \textbf{Average} $\uparrow$ \\
    \midrule
    \rowcolor{gray!30}
    \textit{\textbf{Zero-Shot (Before Fine-Tuning)}}  &  &   &   &   & \\
    Llama-3.1-8B-Instruct & 18.1 & 18.9  &  3.11 &  9.52 &  12.4 \\
    DeepSeek-R1-Distill-Llama-8B & 13.6 & 15.1 & 2.22 & 6.34 & 9.32 \\
    Qwen2.5-7B-Instruct-1M & 18.7 & 18.8 & 6.00 & 10.5 & 13.5 \\
    Qwen3-14B & 15.7 & 19.2 & 3.83 & 8.13 & 11.7 \\
    DeepSeek-R1-Distill-Qwen-14B & 14.0 & 15.5 & 2.40 & 6.67 & 9.64 \\
    Qwen2.5-14B-Instruct-1M & 15.3 & 19.1 & 4.27 & 8.57 & 11.8 \\
    \midrule
    \rowcolor{gray!30}
    \textit{\textbf{Few-Shot (3-shot)}}  &  &   &   &   & \\
    Qwen2.5-7B-Instruct-1M  & 40.7 & 52.7 & 41.0 & 45.7 & 45.0 \\
    \midrule
    \rowcolor{gray!30}
    \textit{\textbf{After Fine-Tuning}}  &  &   &   &   & \\
    Llama-3.1-8B-Instruct & 45.9 & 61.3 & 50.7 & 54.5 &  53.1 \\
    DeepSeek-R1-Distill-Llama-8B & 48.0 & 62.9 & 52.7 & 56.7 & 55.1 \\
    Qwen2.5-7B-Instruct-1M & 49.6 & \textbf{63.8} & \textbf{53.9} & \textbf{58.1} & \textbf{56.4} \\
    Qwen3-14B & \underline{49.9} & 62.9 & 53.1 & 57.4 & 55.8 \\
    DeepSeek-R1-Distill-Qwen-14B & 49.6 & 62.7 & 52.4 & 56.7 & 55.4 \\
    Qwen2.5-14B-Instruct-1M & \textbf{50.8} & \underline{63.5} & \underline{53.3} & \underline{57.6} & \underline{56.3} \\
    \bottomrule
\end{tabular}
\end{table}

\begin{figure}[t]
    \centering
    \includegraphics[width=0.99\linewidth]{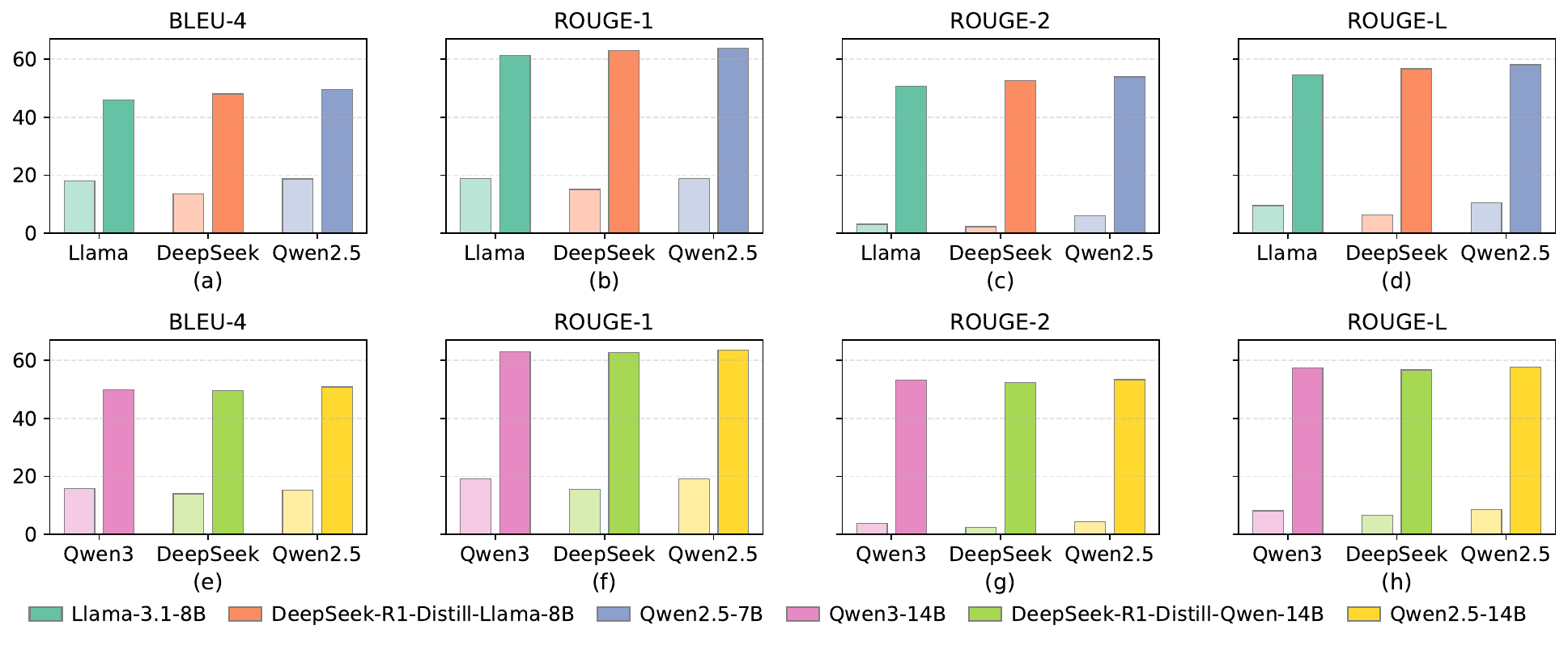}
    \caption{Performance comparison of different LLMs before and after fine-tuning across BLEU and ROUGE metrics. The top row shows 8B/7B-scale models, while the bottom row presents 14B-scale models. For each model, lighter bars denote pre-fine-tuning results and darker bars denote post-fine-tuning performance.}
    \label{fig:overall_performance}
\end{figure}

\subsection{Overall Performance Comparison}

Table~\ref{tab:overall_performance} and Fig.~\ref{fig:overall_performance} present the performance of six LLMs before and after fine-tuning, together with a few-shot prompting baseline on Qwen2.5-7B-Instruct-1M. These models are evaluated on BLEU-4, ROUGE-1, ROUGE-2, and ROUGE-L metrics.
Overall, fine-tuning leads to substantial and consistent performance improvements across all models and evaluation metrics.

Under zero-shot prompting (before fine-tuning), all models exhibit relatively low performance across all metrics, with average scores ranging from 9.32 to 13.5, which indicates that off-the-shelf LLMs struggle to accurately generate structured TMP content. Among the base models, Qwen2.5-7B-Instruct-1M achieves the highest average score of 13.5. Notably, the two DeepSeek-R1-Distill variants rank lowest in this group, suggesting that distillation-based models, while effective for reasoning tasks, do not inherently excel at structured generation tasks without task-specific adaptation.

After fine-tuning, all models show substantial improvements, with average scores increasing by approximately 41--46 points. This dramatic gain underscores the effectiveness of domain-specific fine-tuning for the target task. Among the 7B/8B models, Qwen2.5-7B-Instruct-1M again leads with an average of 56.4, followed by DeepSeek-R1-Distill-Llama-8B at 55.1, while Llama-3.1-8B-Instruct lags behind at 53.1. In the 14B category, Qwen2.5-14B-Instruct-1M achieves the second-best overall score of 56.3, closely trailing its 7B counterpart. Qwen3-14B and DeepSeek-R1-Distill-Qwen-14B also perform competitively, reaching 55.8 and 55.4, respectively. These results suggest that fine-tuning effectively adapts LLMs to the domain-specific structure and terminology of TMP generation.

To examine whether in-context learning can achieve comparable performance without fine-tuning, we further evaluate few-shot prompting on Qwen2.5-7B-Instruct-1M. 
For each target section, three fixed examples randomly sampled from the training set, each consisting of Sections 1–3 and the corresponding target section of a historical TMP, are included in the prompt (\ref{appendix:prompt}).
As shown in Table~\ref{tab:overall_performance}, few-shot prompting substantially improves the average score over zero-shot prompting, from 13.5 to 45.0. This demonstrates that a small number of historical TMP examples can effectively guide LLMs in generating domain-specific content without updating model parameters. However, fine-tuning further improves the average score to 56.4, outperforming few-shot prompting by 11.4 points, with consistent gains of 8.9–12.9 points across all four metrics. These results show that fine-tuning is more effective than few-shot prompting for this task.

Comparing model families, Qwen-based models consistently outperform Llama and DeepSeek counterparts after fine-tuning, indicating stronger adaptability to the task. In addition, increasing model size from 7B/8B to 14B yields only marginal improvements after fine-tuning, as observed in Fig.~\ref{fig:overall_performance}, where performance differences across scales are relatively small. This suggests that task-specific fine-tuning plays a more critical role than model scale in this setting.

Two key observations emerge from these results. First, the performance gap between models narrows considerably after fine-tuning. While pre-fine-tuning scores vary by over 4 points in average, the post-fine-tuning scores of the top five models fall within a narrow 1.3-point range, indicating that fine-tuning acts as a strong equalizer across architectures. Second, scaling from 7B/8B to 14B parameters does not yield consistent improvements: Qwen2.5-7B-Instruct-1M slightly outperforms its 14B counterpart, and no 14B model surpasses the best 7B/8B model. This suggests that, for this particular task, the scaling from 7B/8B to 14B may be insufficient to unlock meaningful gains from increased capacity. A more substantial scaling step, such as moving to 70B-level models, may be necessary to observe clear performance improvements brought by larger parameter counts, which we leave for future investigation. Overall, these results indicate that, under fine-tuning, factors such as model architecture and pre-training data may play a more important role than parameter count when scaling from 7B/8B to 14B.

\subsection{Section-Level Performance Analysis}

\begin{figure}[t]
    \centering
    \includegraphics[width=0.99\linewidth]{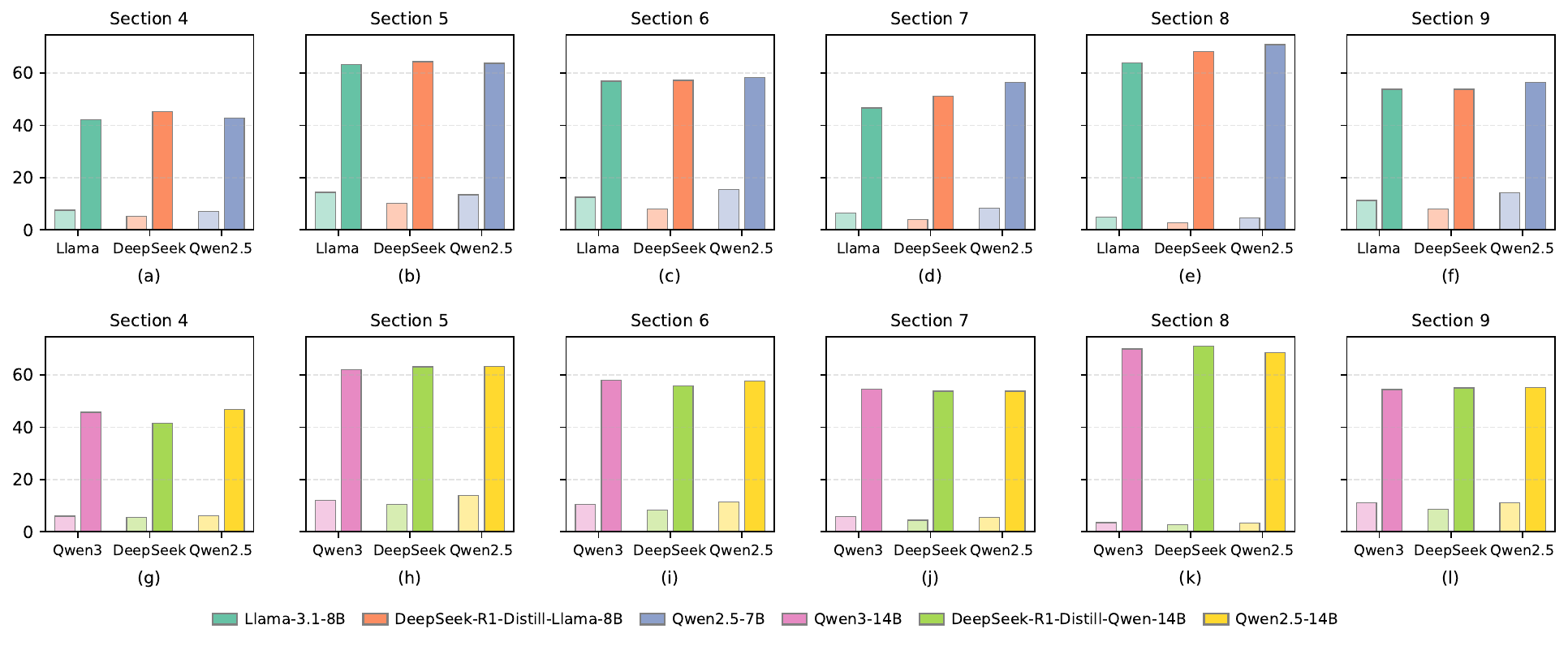}
    \caption{Performance comparison of different LLMs before and after fine-tuning across each section, evaluated using ROUGE-L. The top row shows 8B/7B-scale models, while the bottom row presents 14B-scale models. For each model, lighter bars denote pre-fine-tuning results and darker bars denote post-fine-tuning performance.}
    \label{fig:section_performance}
\end{figure}

To gain deeper insight into model performance across different parts of the TMP document, we evaluate the performance of each model on each target section: Section 4 (Work Zone Strategies), Section 5 (Work Zone Impacts), Section 6 (Traffic Analysis), Section 7 (Public Information Strategies), Section 8 (Incident Management Strategies), and Section 9 (Staging Plans).  Table~\ref{tab:section_performance} reports the BLEU-4 scores before and after fine-tuning, and Fig.~\ref{fig:section_performance} visualizes the corresponding ROUGE-L results.

\begin{table}[t]
\centering
\caption{Performance comparison of different LLMs before and after fine-tuning across each target section, evaluated using BLEU-4. The best results are highlighted in \textbf{bold}, and the second-best results are \underline{underlined}.}
\label{tab:section_performance}
\rowcolors{3}{white}{gray!10}
\resizebox{\textwidth}{!}{
\setlength{\tabcolsep}{4pt}
\begin{tabular}{lcccccc}
    \toprule
     & \multicolumn{6}{c}{\textbf{BLEU-4} $\uparrow$} \\
    \multirow{-2}{*}{\textbf{Model}} & Section 4 & Section 5 & Section 6 & Section 7 & Section 8 & Section 9 \\
    \midrule
    \rowcolor{gray!30}
    \textit{\textbf{Before Fine-Tuning}}  &  &   &   &   &   & \\
    Llama-3.1-8B-Instruct & 13.7 & 32.6 & 21.5 & 13.1 & 7.39 & 20.4 \\
    DeepSeek-R1-Distill-Llama-8B & 11.0 & 22.7 & 17.0 & 8.38 & 3.78 & 19.1 \\
    Qwen2.5-7B-Instruct-1M & 14.7 & 20.1 & 21.7 & 23.5 & 7.12 & 25.1 \\
    Qwen3-14B & 12.0 & 25.1 & 19.2 & 11.8 & 4.73 & 21.3 \\
    DeepSeek-R1-Distill-Qwen-14B & 11.2 & 23.3 & 17.1 & 9.04 & 3.95 & 19.3 \\
    Qwen2.5-14B-Instruct-1M & 11.7 & 24.3 & 18.8 & 12.4 & 4.69 & 20.0 \\
    \midrule
    \rowcolor{gray!30}
    \textit{\textbf{After Fine-Tuning}}  &  &   &   &   &   & \\
    Llama-3.1-8B-Instruct & 33.2 & 57.3 & 50.4 & 40.7 & 41.9 & 52.1 \\
    DeepSeek-R1-Distill-Llama-8B & 37.7 & \textbf{62.3} & 47.8 & 43.8 & 45.1 & 51.4 \\
    Qwen2.5-7B-Instruct-1M & 37.3 & 55.8 & \underline{53.2} & 47.2 & \underline{49.2} & \textbf{55.0} \\
    Qwen3-14B & \underline{38.6} & 57.7 & \textbf{53.8} & 47.8 & 46.7 & \underline{54.9} \\
    DeepSeek-R1-Distill-Qwen-14B & 34.2 & 60.3 & 49.3 & \textbf{48.3} & \textbf{52.9} & 52.8 \\
    Qwen2.5-14B-Instruct-1M & \textbf{41.3} & \underline{61.0} & 52.3 & \underline{48.0} & 47.7 & 54.3 \\
    \bottomrule
\end{tabular}
}
\end{table}

Before fine-tuning, performance varies considerably across sections. Section 5 (Work Zone Impacts) tends to yield the highest BLEU-4 scores for most models (e.g., 32.6 for Llama-3.1-8B-Instruct), while Section 8 (Incident Management Strategies) consistently produces the lowest scores, with all models falling below 7.5. 
This disparity can be largely attributed to the degree to which each section's content aligns with the models' pre-training knowledge. Descriptive and analytical sections such as Work Zone Impacts primarily consist of factual descriptions and analysis, which tend to follow relatively predictable patterns that base models can partially approximate. In contrast, strategy-oriented sections such as Incident Management Strategies and Public Information Strategies require highly domain-specific protocols and structured procedural language that general-purpose models are less likely to have acquired during pre-training.

After fine-tuning, all models exhibit substantial improvements across every section, and the cross-section performance gap narrows notably. Section 8, the most difficult section before fine-tuning, sees the largest absolute gains, with BLEU-4 scores increasing by approximately 35--49 points. This indicates that fine-tuning is particularly effective at addressing the weaknesses of base models on structurally complex and domain-specific sections. Nevertheless, some variation across sections persists: Section 5 (Work Zone Impacts) remains among the highest-scoring sections (reaching up to 62.3), while Section 4 (Work Zone Strategies) tends to yield the lowest post-fine-tuning scores. The relatively lower scores on Work Zone Strategies may stem from the greater diversity and specificity of strategies tailored to individual projects, making this section inherently harder to generate accurately.

Comparing models within each section reveals nuanced differences. Among the 7B/8B models, no single model dominates across all sections: DeepSeek-R1-Distill-Llama-8B achieves the highest BLEU-4 on Section 5 (62.3), while Qwen2.5-7B-Instruct-1M leads on Sections 6, 7, 8, and 9. In the 14B category, Qwen2.5-14B-Instruct-1M achieves the best score on Section 4 (41.3), while DeepSeek-R1-Distill-Qwen-14B excels on Section 8 (52.9). The ROUGE-L results shown in Fig.~\ref{fig:section_performance} largely corroborate these trends, with similar relative rankings observed across models and sections.

These per-section results highlight two important findings. First, fine-tuning not only improves overall performance but also reduces the performance imbalance across sections, enabling models to handle both standardized content (e.g., Work Zone Impacts) and highly specialized content (e.g., Incident Management Strategies) more uniformly. Second, section-specific difficulty persists even after fine-tuning, indicating that generation performance depends not only on model capacity but also on the inherent characteristics of the target content.

\subsection{Strategy-Level Performance Analysis}
\label{sec:strategy_performance}

\begin{table}[t]
\caption{Strategy-level recall and precision of fine-tuned models on Section~4 (Work Zone Strategies), Section~7 (Public Information Strategies), and Section~8 (Incident Management Strategies). The best results are highlighted in \textbf{bold}, and the second-best results are \underline{underlined}.}
\label{tab:strategy_performance}
\centering
\resizebox{\textwidth}{!}{
\setlength{\tabcolsep}{4pt}
\rowcolors{3}{gray!10}{white}
\begin{tabular}{l|cccc|cccc}
    \toprule
    \rowcolor{gray!30}
     & \multicolumn{4}{c|}{\textbf{Recall (\%)} $\uparrow$} & \multicolumn{4}{c}{\textbf{Precision (\%)} $\uparrow$ } \\
    \rowcolor{gray!30}
    \multirow{-2}{*}{\textbf{Model}} & Section 4 & Section 7 & Section 8 & Average & Section 4 & Section 7 & Section 8 & Average \\
    \midrule
    Llama-3.1-8B-Instruct & 36.70 & 51.59 & 65.00 & 51.09 & 29.82 & 44.97 & 49.27 & 41.34 \\
    DeepSeek-R1-Distill-Llama-8B & 34.84 & 58.19 & 47.92 & 46.94 & 27.59 & 47.70 & 47.50 & 40.90 \\
    Qwen2.5-7B-Instruct-1M & 37.90 & 62.50 & 54.79 & \underline{51.68} & 29.66 & 54.60 & 50.94 & 45.02 \\
    Qwen3-14B & 38.31 & 57.65 & 59.38 & \textbf{51.75} & 32.70 & 53.67 & 56.25 & \textbf{47.51} \\
    DeepSeek-R1-Distill-Qwen-14B & 36.94 & 55.43 & 59.38 & 50.56 & 25.51 & 50.52 & 55.00 & 43.65 \\
    Qwen2.5-14B-Instruct-1M & 40.45 & 58.03 & 55.00 & 51.13 & 31.67 & 53.63 & 50.62 & \underline{45.27} \\
    \bottomrule
\end{tabular}
}
\end{table}

To further examine model performance beyond conventional text similarity metrics, we evaluate generation quality for Section 4 (Work Zone Strategies), Section 7 (Public Information Strategies), and Section 8 (Incident Management Strategies) using precision and recall at the strategy level. This domain-specific evaluation is particularly important for TMP preparation, as generated content may exhibit high textual similarity to a historical TMP while omitting important traffic management strategies or introducing additional ones. By evaluating individual strategies rather than textual overlap, these metrics provide complementary insights into the capabilities and limitations of LLMs in TMP preparation.

Table~\ref{tab:strategy_performance} shows that average recall consistently exceeds average precision across all models, which suggests a tendency toward strategy over-generation. Meanwhile, the relatively modest recall values indicate that a substantial proportion of reference strategies are still omitted, which highlights the difficulty of project-specific strategy selection even after fine-tuning.
Across sections, a consistent pattern emerges. Section~8 (Incident Management Strategies) yields the highest recall for most models (e.g., 65.00 for Llama-3.1-8B-Instruct), suggesting that incident management strategies are relatively standardized and easier to cover comprehensively. In contrast, Section~4 (Work Zone Strategies) exhibits the lowest recall across all models (generally below 41\%).
This pattern is consistent with the per-section ROUGE-L findings reported in  Table~\ref{tab:section_performance} and Fig.~\ref{fig:section_performance}, where Section 4 also exhibits the lowest post-fine-tuning scores. 
The relatively lower performance on Work Zone Strategies likely reflects the greater diversity of project-specific strategies in this section, which makes exact recovery more challenging.

Comparing models, Qwen3-14B achieves the best overall performance, with the highest average recall (51.75\%) and precision (47.51\%), indicating a better balance between coverage and accuracy. Among smaller models, Qwen2.5-7B-Instruct-1M performs competitively, achieving the second-best recall (51.68\%) and strong precision (45.02\%), suggesting that architecture and training data can outweigh parameter scaling at this range. In contrast, the distilled DeepSeek models show relatively lower precision, implying that while reasoning-oriented distillation may help coverage in some cases (e.g., Section~7 recall), it does not consistently improve the accuracy of selected strategies.

These results reveal a trade-off between coverage and precision in strategy generation, which highlights limitations that may not be adequately captured by conventional text similarity metrics alone. However, strategy-level precision and recall measure agreement with historical TMPs rather than engineering appropriateness. Strategies that are not present in the reference TMP are not necessarily inappropriate, as multiple valid strategies may exist for a given project. Therefore, the generated strategies should be treated as preliminary recommendations subject to practitioner review rather than definitive engineering decisions. Future work could explore retrieval-augmented or constrained decoding approaches to improve strategy selection, as well as practitioner-based evaluations to assess the technical correctness and practical usefulness of generated strategies.

\subsection{Qualitative Case Study}

\begin{table}[t]
\centering
\caption{Case 1: Strategy-level comparison between generated and ground-truth TMP strategies for Section~4 (Work Zone Strategies), with strategies grouped into matched and unmatched categories.}
\label{tab:case1}
\rowcolors{3}{gray!10}{white}
\resizebox{\textwidth}{!}{
\begin{tabular}{p{0.5\textwidth} p{0.5\textwidth}}
    \toprule
    \textbf{Ground Truth Strategies} & \textbf{Generated Strategies} \\
    \midrule
    \rowcolor{gray!30}
    \textit{\textbf{Matched Strategies}}  &  \\
    Continuous Full Closure (for a project phase/stage or the entire project) & Continuous Full Closure (for a project phase/stage or the entire project)\\
    Detour Route & Detour Route \\
    Portable Changeable Message Signs (PCMS) & Portable Changeable Message Signs (PCMS) \\
    \midrule
    \rowcolor{gray!30}
    \textit{\textbf{Unmatched Strategies}}  &  \\
     & Coordination with Other Projects \\
    \bottomrule
\end{tabular}
}
\end{table}

\begin{table}[t]
\centering
\caption{Case 2: Strategy-level comparison between generated and ground-truth TMP strategies for Section~4 (Work Zone Strategies), with strategies grouped into matched and unmatched categories.}
\label{tab:case2}
\rowcolors{3}{gray!10}{white}
\resizebox{\textwidth}{!}{
\begin{tabular}{p{0.5\textwidth} p{0.5\textwidth}}
    \toprule
    \textbf{Ground Truth Strategies} & \textbf{Generated Strategies} \\
    \midrule
    \rowcolor{gray!30}
    \textit{\textbf{Matched Strategies}}  &  \\
    Construction phasing/staging & Construction phasing/staging\\
    Lane closures & Lane closures \\
    Ramp Closures & Ramp Closures \\
    Night work & Night work \\
    Lane Rental & Lane Rental \\
    Temporary Pavement Markings (does not include chevrons or orange pavement markings) & Temporary Pavement Markings (does not include chevrons or orange pavement markings) \\
    Temporary Concrete Barrier & Temporary Concrete Barrier \\
    Crash Cushions & Crash Cushions \\
    Channelizing Devices & Channelizing Devices \\
    Temporary Speed Limit Reduction /Temporary Speed Zone Declaration & Temporary Speed Limit Reduction /Temporary Speed Zone Declaration \\
    Detour Route & Detour Route \\
    Temporary Traffic Control Signs & Temporary Traffic Control Signs \\
    Portable Changeable Message Signs (PCMS) & Portable Changeable Message Signs (PCMS) \\
    Arrow Panels/Board & Arrow Panels/Board \\
    \midrule
    \rowcolor{gray!30}
    \textit{\textbf{Unmatched Strategies}}  &  \\
    Reduced Lane Widths & Reduced Shoulder Width \\
    Alternate Route & Lane Shift to Shoulder/Median \\
     & Shoulder Closure \\
     & Work hour restrictions for peak travel \\
     & Coordination with Other Projects \\
    \bottomrule
\end{tabular}
}
\end{table}

To further understand model performance at the strategy level, we conduct qualitative case studies on Section~4 (Work Zone Strategies) using the fine-tuned Qwen2.5-7B-Instruct-1M model. Tables~\ref{tab:case1} and~\ref{tab:case2} present two representative examples, where strategies are grouped into matched and unmatched categories.

As shown in Table~\ref{tab:case1}, case 1 involves a relatively simple project with only three ground-truth strategies. The model successfully identifies all three and achieves perfect recall. However, it also generates one additional strategy, Coordination with Other Projects, that does not appear in the ground truth. Although this strategy is generally relevant to work zone planning, it is not required for the specific project, and its necessity for the specific project cannot be established from the available information. This example illustrates a common pattern where models achieve strong recall but may slightly over-generate, resulting in reduced precision.

 Table~\ref{tab:case2} presents a more complex case with a larger set of strategies. The model correctly recovers 14 of the 16 ground-truth strategies, missing only Reduced Lane Widths and Alternate Route. At the same time, it introduces five strategies not present in the ground truth: Reduced Shoulder Width, Lane Shift to Shoulder/Median, Shoulder Closure, Work Hour Restrictions for Peak Travel, and Coordination with Other Projects. Interestingly, some of the unmatched strategies on both sides are semantically related. For example, the ground truth specifies Reduced Lane Widths while the model generates Reduced Shoulder Width, suggesting that the model captures the general intent of geometric adjustments but selects a different specific measure. This pattern indicates that certain errors may stem from confusion among closely related strategy types rather than from a fundamental misunderstanding of the project context.

Overall, these case studies reveal a consistent trend: models are effective at identifying a broad set of relevant strategies, leading to high recall, but they tend to over-generate and include loosely related or generalized strategies, which reduces precision. These observations are consistent with the quantitative results reported in Sec.~\ref{sec:strategy_performance}, where recall is generally higher than precision. They also highlight the inherent difficulty of Section~4, where strategy selection is highly project-specific and admits multiple plausible alternatives, making precise matching more challenging.

\begin{table}[t]
\centering
\caption{Average inference time (mean $\pm$ standard deviation, in seconds) for each fine-tuned model across TMP sections (Sections 4--9), along with the total time for generating all six sections.}
\label{tab:inference_time}
\rowcolors{1}{gray!10}{white}
\resizebox{\textwidth}{!}{
\setlength{\tabcolsep}{4pt}
\begin{tabular}{lccccccc}
    \toprule
    \rowcolor{gray!30}
     & \multicolumn{7}{c}{\textbf{Inference Time (s)} $\downarrow$} \\
    \rowcolor{gray!30}
    \multirow{-2}{*}{\textbf{Model}} & Section 4 & Section 5 & Section 6 & Section 7 & Section 8 & Section 9 & Total \\
    \midrule
    Llama-3.1-8B-Instruct & 11.8 {\tiny $\pm$ 6.93} & 9.00 {\tiny $\pm$ 3.41} & 12.0 {\tiny $\pm$ 8.52} & 6.84 {\tiny $\pm$ 5.78} & 2.80 {\tiny $\pm$ 1.43} & 13.6 {\tiny $\pm$ 8.90} & 55.8 {\tiny $\pm$ 22.1} \\
    DeepSeek-R1-Distill-Llama-8B & 9.38 {\tiny $\pm$ 7.40} & 9.50 {\tiny $\pm$ 3.48} & 11.6 {\tiny $\pm$ 9.15} & 6.56 {\tiny $\pm$ 4.21} & 2.36 {\tiny $\pm$ 1.21} & 12.4 {\tiny $\pm$ 8.21} & 51.6 {\tiny $\pm$ 22.7} \\
    Qwen2.5-7B-Instruct-1M & 10.2 {\tiny $\pm$ 7.69} & 9.90 {\tiny $\pm$ 5.10} & 10.1 {\tiny $\pm$ 7.07} & 7.52 {\tiny $\pm$ 4.92} & 8.65 {\tiny $\pm$ 7.24} & 11.7 {\tiny $\pm$ 7.94} & 57.7 {\tiny $\pm$ 24.3} \\
    Qwen3-14B & 23.2 {\tiny $\pm$ 16.8} & 17.1 {\tiny $\pm$ 8.64} & 22.7 {\tiny $\pm$ 16.2} & 10.9 {\tiny $\pm$ 8.40} & 5.87 {\tiny $\pm$ 3.66} & 23.8 {\tiny $\pm$ 15.4} & 102.9 {\tiny $\pm$ 48.6} \\
    DeepSeek-R1-Distill-Qwen-14B & 19.7 {\tiny $\pm$ 12.9} & 18.8 {\tiny $\pm$ 10.0} & 22.5 {\tiny $\pm$ 15.6} & 10.8 {\tiny $\pm$ 5.79} & 5.18 {\tiny $\pm$ 5.64} & 23.9 {\tiny $\pm$ 15.6} & 100.7 {\tiny $\pm$ 48.5} \\
    Qwen2.5-14B-Instruct-1M & 16.7 {\tiny $\pm$ 11.7} & 18.3 {\tiny $\pm$ 11.0} & 23.3 {\tiny $\pm$ 17.5} & 7.19 {\tiny $\pm$ 3.68} & 4.11 {\tiny $\pm$ 2.02} & 22.2 {\tiny $\pm$ 15.0} & 91.2 {\tiny $\pm$ 15.0} \\
    \bottomrule
\end{tabular}
}
\end{table}

\subsection{Inference Efficiency Analysis}

To assess the computational efficiency of the fine-tuned models, we measure the average inference time required to generate each target section on a single NVIDIA RTX A6000 GPU.
Table~\ref{tab:inference_time} reports the mean and standard deviation of inference time across all test samples for each model and section.
Among the 7B/8B models, all three complete the generation of all six operational TMP sections (Sections 4--9) in under one minute. In contrast, the 14B models require approximately twice as long, with total inference times ranging from 91 to 103 seconds.
Across sections, inference time generally correlates with output length, as characterized in Section~\ref{sec:data_stats}. Sections 5, 6, and 9, which produce longer outputs (averaging 217--238 words), consistently require more time across all models. In contrast, Section 8 (Incident Management Strategies), the shortest section with an average of 37 words, is generated most quickly, often in under 6 seconds.

From a practical perspective, even the largest evaluated model can generate a complete set of operational TMP sections in under two minutes on average, which demonstrates the computational feasibility of LLM-assisted TMP drafting. Rather than preparing each operational section from scratch, practitioners could use the generated content as an initial draft for subsequent review and refinement. This workflow has the potential to reduce the time and effort required for TMP preparation.

\subsection{Error Analysis and Limitations}

\begin{table}[t]
\centering
\caption{Error case: comparison between generated and ground-truth Section 4 strategies, justifications, and cost estimates produced by fine-tuned Qwen2.5-7B-Instruct-1M.}
\label{tab:error}

\begin{minipage}[t]{0.48\textwidth}
\centering
\small{(a) Ground Truth}

\vspace{0.5em}
\small

\resizebox{\textwidth}{!}{
\rowcolors{1}{white}{gray!10}
\begin{tabular}{p{2.8cm} p{4.5cm} p{0.8cm}}
\toprule
\rowcolor{gray!30} 
\textbf{Strategy} & \textbf{Justification/Comment} & \textbf{Cost} \\
\midrule
Shoulder Closure
& 
& \$500 \\

Flagging Operation/One-lane, Two-Way Operation
&  
& \$1000 \\

Temporary Traffic Control Signs
&
& \$500 \\

Fixed Message Signs
&
& \$500 \\
\bottomrule
\end{tabular}
}
\end{minipage}
\hfill
\begin{minipage}[t]{0.48\textwidth}
\centering
\small{(b) Generation Results}

\vspace{0.5em}
\small
\resizebox{\textwidth}{!}{
\rowcolors{1}{white}{gray!10}
\begin{tabular}{p{2.8cm} p{4.5cm} p{0.8cm}}
\toprule
\rowcolor{gray!30}
\textbf{Strategy} & \textbf{Justification/Comment} & \textbf{Cost} \\
\midrule

Construction phasing/staging
& Work will be completed in 3 stages. See attached ``Project Staging" sheet for details.
& \$0 \\

Continuous Full Closure (for a project phase/stage or the entire project)
& Culvert replacements will require full closure of the road during construction. A full closure will also be required for a short period of time to line culverts and replace culverts. See attached ``Traffic Control" sheet for details.
& \$50000 \\

Flagging Operation /One-lane, Two-Way Operation
& Used during milling and paving operations. See attached ``Traffic Control" sheet for details.
& \$0 \\

Temporary Traffic Control Signs
& Temporary traffic control signs will be placed as shown in the traffic control sheet.
& \$0 \\

Portable Changeable Message Signs (PCMS)
& PCMS will be used 1 week prior to the start of construction to alert drivers of the upcoming project.
& \$2000 \\

Coordination with Other Projects
& No conflicts expected.
& \$0 \\

\bottomrule
\end{tabular}
}
\end{minipage}

\end{table}

To better understand the limitations of the fine-tuned models beyond aggregate evaluation metrics, we examine a representative error case generated by Qwen2.5-7B-Instruct-1M for Section~4 (Work Zone Strategies), as shown in Table~\ref{tab:error}. In this example, the ground truth contains four strategies, while the model correctly recovers only two of them (\emph{Flagging Operation/One-lane, Two-Way Operation} and \emph{Temporary Traffic Control Signs}). At the same time, it introduces four additional strategies that are not present in the ground truth. This example illustrates both omission and over-generation errors.
Beyond strategy selection, the generated outputs in Table~\ref{tab:error} reveal important limitations in two other dimensions of TMP content: justification/comment and cost.

For justification/comment, the model exhibits a tendency to generate additional explanatory text even when no justification is provided in the ground truth. As shown in Table~\ref{tab:error}(a), all justification/comment fields are empty in the ground truth TMP. However, the model generates detailed descriptions for every predicted strategy, including statements such as ``Work will be completed in 3 stages'' and ``PCMS will be used 1 week prior to the start of construction.'' While these comments are generally plausible and consistent with common TMP practices, they are not supported by the available input information and therefore cannot be verified against the ground truth document. This behavior suggests that the model has learned common TMP writing patterns from the training data and tends to fill missing fields with generic but plausible content, resulting in another form of over-generation.

Cost estimation presents another limitation. Most generated strategies are assigned either \$0 or inaccurate values (e.g., \$2{,}000 for PCMS and \$50{,}000 for Continuous Full Closure), while the ground-truth costs range from only \$500 to \$1{,}000. This behavior is largely expected because the input to the model is primarily derived from Sections~1--3 of the TMP, which do not contain sufficient information for reliable cost estimation. In practice, costs depend on project-specific factors such as project scope, duration, traffic control requirements, material prices, and local labor conditions. Consequently, accurate cost prediction remains beyond the capability of the current text-generation framework and would likely require integration with external cost databases.

\section{Conclusion and Future Work}

This paper presents an LLM-assisted framework for automating the preparation of TMP, using the WisDOT WisTMP system as the application context. The proposed approach constructs a domain-specific dataset from 843 historical TMP documents, converting them into 5,048 structured question–answer pairs through a pipeline involving PDF-to-Markdown conversion, multi-stage data cleaning, and section-based QA pair generation. Six open-source LLMs from three model families (LLaMA, Qwen, and DeepSeek) at two parameter scales (7B/8B and 14B) are fine-tuned using LoRA and systematically evaluated for TMP content generation.

The experimental results yield several key findings. First, fine-tuning produces substantial performance gains across all models, with average scores improving by approximately 41--46 points on BLEU-4 and ROUGE metrics, demonstrating that domain-specific adaptation is essential for structured TMP generation. 
Second, few-shot prompting with three historical TMP examples substantially improves performance without updating model parameters, raising the average score of Qwen2.5-7B-Instruct-1M from 13.5 under zero-shot prompting to 45.0. However, fine-tuning achieves higher scores on all four metrics, with an average of 56.4, which indicates that parameter-level adaptation provides additional benefits under the evaluated settings.
Second, Qwen-based models consistently achieve the strongest performance after fine-tuning, with Qwen2.5-7B-Instruct-1M attaining the highest overall average score of 56.4 among all evaluated models. Third, scaling from 7B/8B to 14B parameters yields only marginal improvements, suggesting that task-specific fine-tuning plays a more decisive role than model scale for this application. Fourth, per-section analysis reveals that fine-tuning is particularly effective at improving performance on domain-specific sections such as Incident Management Strategies, which exhibit the largest absolute gains. Finally, strategy-level evaluation shows that fine-tuned models achieve reasonable recall in recovering ground-truth strategies but tend to over-generate, resulting in a consistent gap between recall and precision. 

Despite these promising results, several limitations and directions for future work remain. First, the current framework generates each TMP section independently, without modeling inter-section dependencies. Future work could explore multi-section joint generation or hierarchical generation approaches that capture the logical relationships among different parts of a TMP. 
Second, as revealed by the error analysis, the model struggles with generating project-specific justifications and accurate cost estimates. Integrating retrieval-augmented generation (RAG) with external knowledge sources, such as historical cost databases, regional construction pricing indices, and project-specific geospatial data, could improve the specificity and accuracy of generated content. 
Third, although the current evaluation incorporates both text-level and domain-specific strategy-level metrics, agreement with historical TMPs does not necessarily guarantee the technical correctness, regulatory compliance, or practical usefulness of generated content. Future studies should include human-in-the-loop evaluations with transportation practitioners to assess these aspects, as well as the actual time savings of LLM-assisted TMP preparation in real-world workflows.
Fourth, the dataset is dominated by Type 1 and Type 2 projects from Wisconsin. Future work should evaluate the framework on more complex projects and on TMPs from other states, which would require fine-tuning with state-specific data.
Fifth, while scaling from 7B/8B to 14B did not yield significant gains, exploring substantially larger models (e.g., 70B parameters) or more advanced fine-tuning strategies such as reinforcement learning may further improve generation quality. 
Finally, incorporating multimodal inputs such as roadway images, traffic flow diagrams, and geographic information system (GIS) data could enable the model to better understand project-specific spatial and geometric conditions, further enhancing the relevance and accuracy of generated TMP content.

\section*{Declaration of Competing Interest}
The authors declare that they have no known competing financial interests or personal relationships that could have appeared to influence the work reported in this paper.

\section*{Acknowledgment}
This work was supported by the University of Wisconsin-Madison's Center for Connected and Automated Transportation (CCAT), a part of the larger CCAT consortium, a USDOT Region 5 University Transportation Center funded by the U.S. Department of Transportation, Award \#69A3552348305. The contents of this paper reflect the views of the authors, who are responsible for the facts and the accuracy of the data presented herein, and do not necessarily reflect the official views or policies of the sponsoring organization.

The authors would also like to acknowledge the Bureau of Traffic Operations (BTO) at the Wisconsin Department of Transportation (WisDOT) for their valuable data, support, and domain expertise.
In addition, this work was supported by grants from NVIDIA and utilized NVIDIA RTX PRO 6000 Blackwell Max-Q Workstation Edition GPUs.

\appendix

\section{TMP Example for WisTMP}\label{appendix:tmp_example}

\begin{figure}[h]
    \centering
    \includegraphics[width=0.9\textwidth]{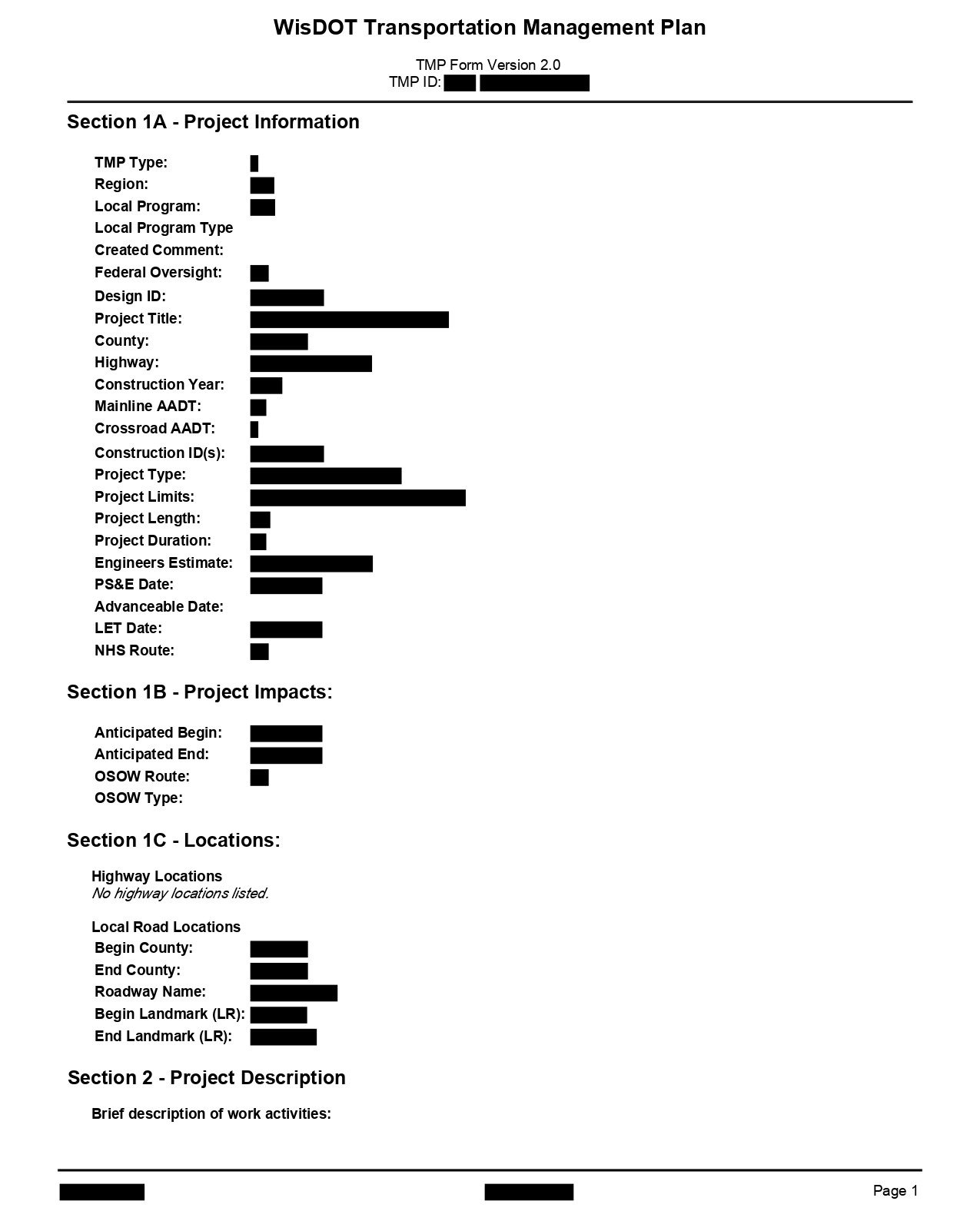}
    \caption{Example TMP (Page 1).}
    \label{fig:tmp_page1}
\end{figure}

\begin{figure}[p]
    \centering
    \includegraphics[width=\textwidth]{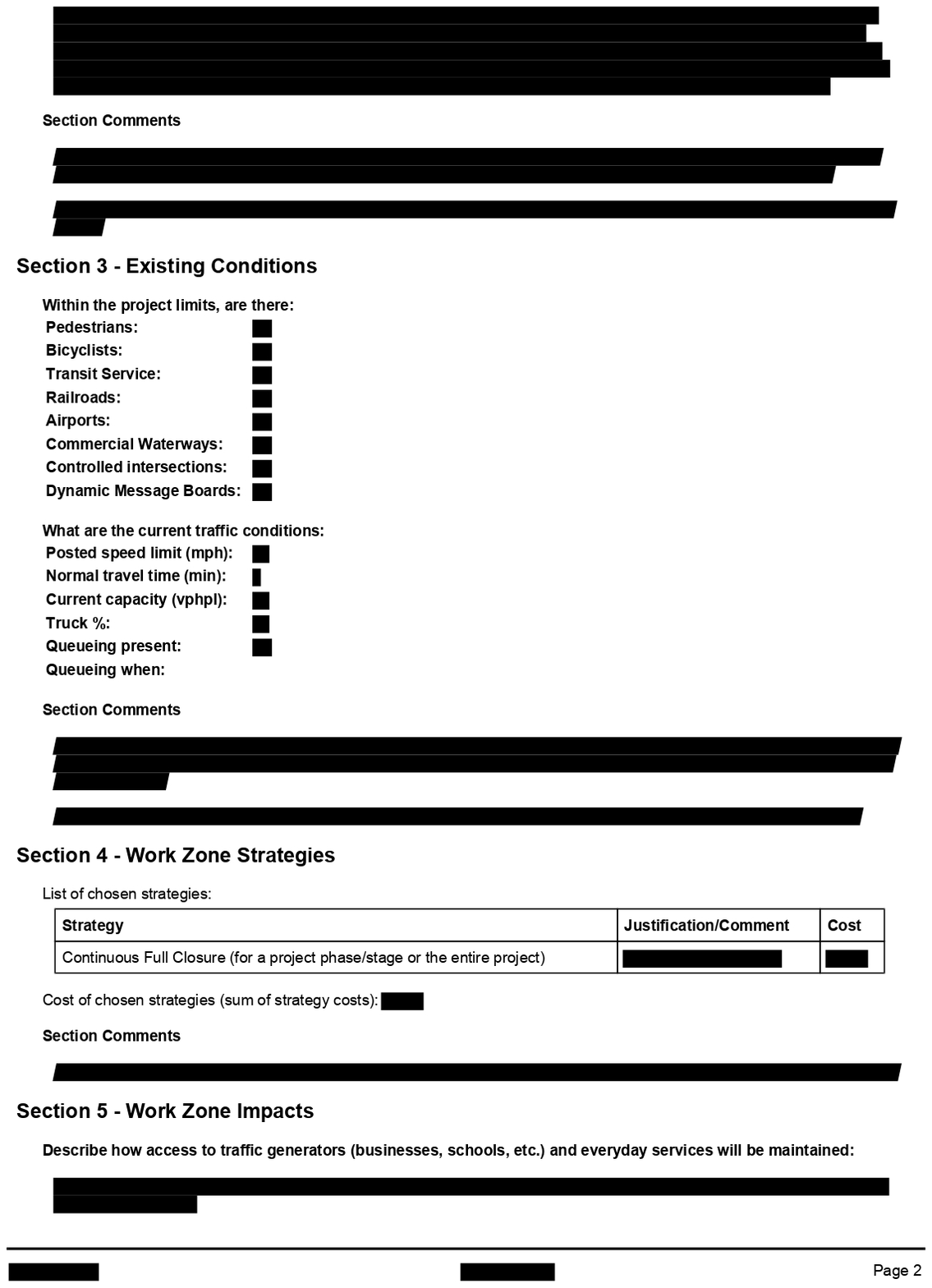}
    \caption{Example TMP (Page 2).}
    \label{fig:tmp_page2}
\end{figure}

\begin{figure}[p]
    \centering
    \includegraphics[width=\textwidth]{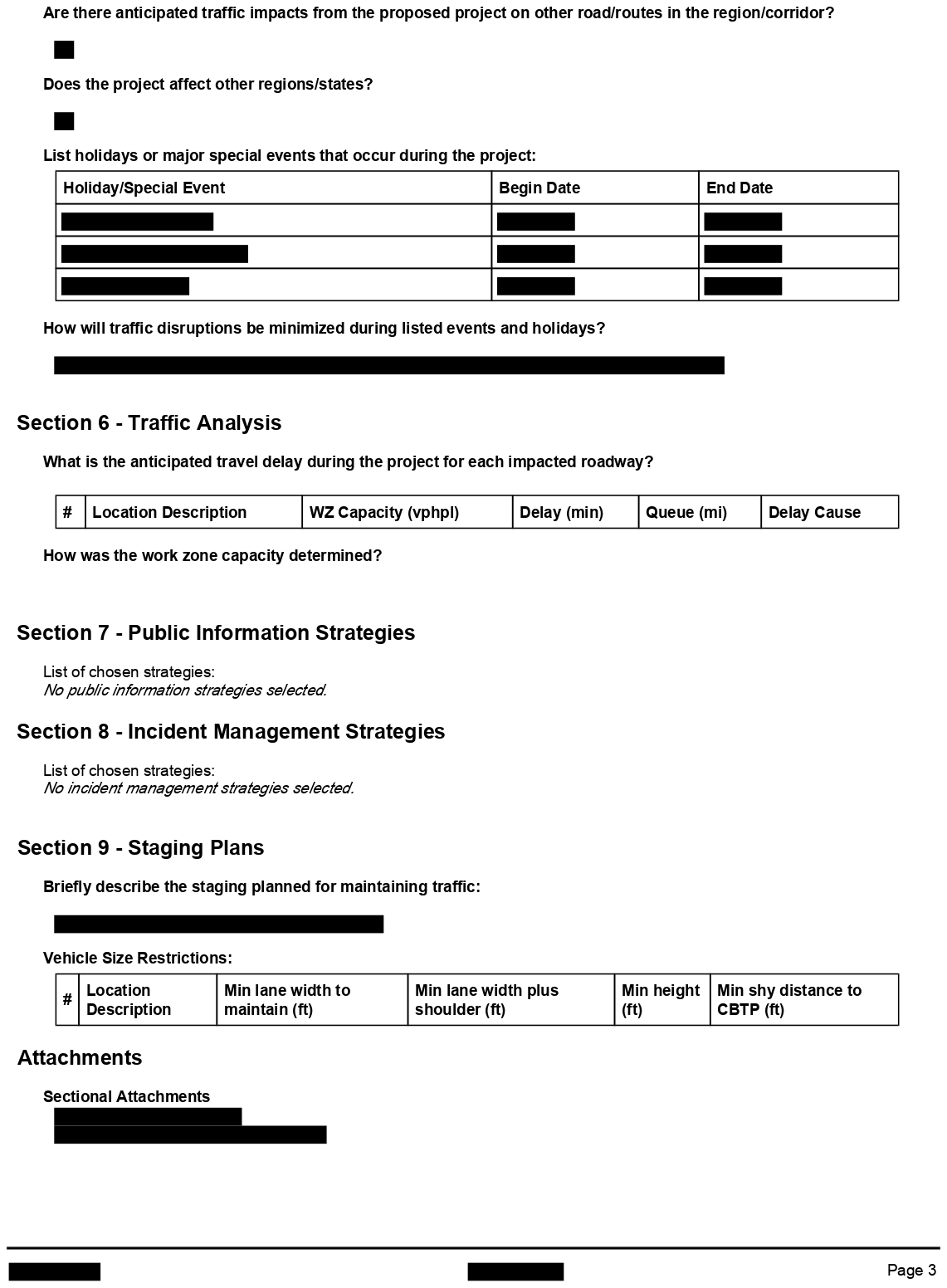}
    \caption{Example TMP (Page 3).}
    \label{fig:tmp_page3}
\end{figure}

As shown in Figs.~\ref{fig:tmp_page1}--\ref{fig:tmp_page3}, this appendix provides a representative TMP example from the WisTMP system, with all sensitive and identifying information removed. 

\newpage
\section{Prompt Templates for TMP Content Generation}
\label{appendix:prompt}

This appendix presents the prompt templates used for TMP content generation. Table~\ref{tab:standard_prompt} presents the standard prompt template used for zero-shot prompting and fine-tuned LLMs, while Table~\ref{tab:fewshot_prompt} presents the template used for few-shot prompting. In both templates, the target section $k \in \{4, 5, 6, 7, 8, 9\}$ is specified according to the generation task, and the project context is derived from Sections 1--3 of the TMP document.

All placeholders in these templates are replaced with the corresponding TMP content during inference. For few-shot prompting, demonstrations are exclusively drawn from the training set to avoid data leakage.

\begin{table*}[h]
{
\centering
\caption{Standard prompt template used for zero-shot prompting 
and fine-tuned LLMs. The model receives a task instruction and 
project context from Sections 1--3 and generates the specified 
operational section.}
\label{tab:standard_prompt}

\begin{tcolorbox}[halign=left]

\textbf{Task Instruction}

You are a helpful assistant.

Filling \texttt{Section \textless{}SECTION\_ID\textgreater{} - \textless{}SECTION\_NAME\textgreater{}}
\\[6pt]

\textbf{Project Context}

Section 1: Project Information

\texttt{<SECTION\_1\_CONTENT>}

Section 2: Project Description

\texttt{<SECTION\_2\_CONTENT>}

Section 3: Existing Conditions

\texttt{<SECTION\_3\_CONTENT>}

\end{tcolorbox}
}
\end{table*}

\begin{table*}[h]
{
\centering
\caption{Few-shot prompt template used for TMP content generation. The model receives a task instruction, three input--output demonstrations drawn from the training set, and the project context of the target TMP. All demonstrations correspond to the same target section as the current generation task.}
\label{tab:fewshot_prompt}

\begin{tcolorbox}[halign=left]

\textbf{Task Instruction}
You assist engineers in preparing WisDOT Transportation Management Plans. Generate the requested section from the supplied project context. Output only that section in Markdown.

Filling \texttt{Section \textless{}SECTION\_ID\textgreater{} - \textless{}SECTION\_NAME\textgreater{}}
\\[6pt]

\textbf{Demonstration 1}

Input: \texttt{<EXAMPLE\_1\_SECTIONS\_1\_TO\_3>}

Output: \texttt{<EXAMPLE\_1\_TARGET\_SECTION>}
\\[6pt]

\textbf{Demonstration 2}

Input: \texttt{<EXAMPLE\_2\_SECTIONS\_1\_TO\_3>}

Output: \texttt{<EXAMPLE\_2\_TARGET\_SECTION>}
\\[6pt]

\textbf{Demonstration 3}

Input: \texttt{<EXAMPLE\_3\_SECTIONS\_1\_TO\_3>}

Output: \texttt{<EXAMPLE\_3\_TARGET\_SECTION>}
\\[6pt]

\textbf{Target Project}

Input: \texttt{<TARGET\_SECTIONS\_1\_TO\_3>}

Output:

\end{tcolorbox}
}
\end{table*}

\newpage
\bibliographystyle{elsarticle-harv}
\biboptions{authoryear}
\bibliography{reference}

\end{document}